\documentclass[journal,twoside,web]{ieeecolor}
\usepackage{generic}
\usepackage{cite}
\usepackage{amsmath,amssymb,amsfonts}
\usepackage{algorithmic}
\usepackage{graphicx}
\usepackage{algorithm,algorithmic}
\usepackage{hyperref}
\hypersetup{hidelinks=true}
\usepackage{textcomp}
\usepackage{booktabs}
\usepackage{multirow}
\usepackage{threeparttable}
\usepackage{bm} 

\def\BibTeX{{\rm B\kern-.05em{\sc i\kern-.025em b}\kern-.08em
    T\kern-.1667em\lower.7ex\hbox{E}\kern-.125emX}}
\begin{document}
\title{TSP-OCS: A Time-Series Prediction for Optimal Camera Selection in Multi-Viewpoint Surgical Video Analysis}
\author{Xinyu Liu, Xiaoguang Lin, Xiang Liu,Yong Yang, Hongqian Wang,Qilong Sun 
\thanks{This manuscript was accepted on October 31, 2024; this work was supported by the First Affiliated Hospital of Army Medical University and the Chongqing Institute of Green Intelligent Technology, Chinese Academy of Sciences. (Corresponding author: Xiaoguang Lin)}
\thanks{Xinyu Liu is affiliated with Chongqing Institute of Green and Intelligent Technology, Chinese Academy of Sciences, and Chongqing School, University of Chinese Academy of Sciences, Chongqing, 400714, China(e-mail: liuxinyu233@mails.ucas.ac.cn).}
\thanks{Xiaoguang Lin, Yong Yang, and Qilong Sun are affiliated with Chongqing Institute of Green and Intelligent Technology, Chinese Academy of Sciences, Chongqing, 400714, China (e-mail: lxg@cigit.ac.cn; yangyong@cigit.ac.cn; sunqilong@cigit.ac.cn).}
\thanks{Xiang Liu and Hongqian Wang are affiliated with Southwest Hospital, Third Military Medical University, Chongqing, 400038, China(e-mail: lewis.liuxiang@outlook.com; wanghongqianseu@163.com).}
}
\maketitle
\begin{abstract}
Recording the open surgery process is essential for educational and medical evaluation purposes; however, traditional single-camera methods often face challenges such as occlusions caused by the surgeon's head and body, as well as limitations due to fixed camera angles, which reduce comprehensibility of the video content. This study addresses these limitations by employing a multi-viewpoint camera recording system, capturing the surgical procedure from six different angles to mitigate occlusions. We propose a fully supervised learning-based time series prediction method to choose the best shot sequences from multiple simultaneously recorded video streams, ensuring optimal viewpoints at each moment. Our time series prediction model forecasts future camera selections by extracting and fusing visual and semantic features from surgical videos using pre-trained models. These features are processed by a temporal prediction network with TimeBlocks to capture sequential dependencies. A linear embedding layer reduces dimensionality, and a Softmax classifier selects the optimal camera view based on the highest probability. In our experiments, we created five groups of open thyroidectomy videos, each with simultaneous recordings from six different angles. The results demonstrate that our method achieves competitive accuracy compared to traditional supervised methods, even when predicting over longer time horizons. Furthermore, our approach outperforms state-of-the-art time series prediction techniques on our dataset. This manuscript makes a unique contribution by presenting an innovative framework that advances surgical video analysis techniques, with significant implications for improving surgical education and patient safety.
\end{abstract}


\section{Introduction}
\IEEEPARstart{R}{ecording} surgery scenes preserves crucial surgical information. As artificial intelligence technology continues to advance, the application of open surgical scene recording has expanded beyond traditional educational and medical sharing \cite{sadri2013video,matsumoto2013digital}, making it possible in tasks such as surgical scene understanding, event detection, and data-driven decision support \cite{acosta2022multimodal}. However, the complexity of open surgical scenes is a big challenge to the recording process. Traditional single-camera recording methods may result in significant data loss, since the bodies of doctors and nurses inevitably block the surgical image in the surgical area, and the single camera has a high risk of instability.

Liu et al. \cite{bib1} and Shimizu et al. \cite{bib2} both explored multi-viewpoint camera recording techniques to mitigate visual information loss in surgical videos caused by object occlusions. Multiple cameras were mounted on the shadowless lamp at different angles to comprehensively capture the surgical scene. The shadowless lamp provides uniform illumination during surgery, ensuring critical surgical areas are well-lit and captured by multiple cameras. Moreover, the shadowless lamp's design minimizes shadows in the surgical field, significantly facilitating image processing and enhancing visual clarity. However, the data volume generated by multi-viewpoint camera setups multiplies significantly. This results in a significant amount of invalid and redundant video data, complicating processing and comprehension of the information. A multi-viewpoint camera switching algorithm enables the selection and output of the optimal view from multiple cameras, enhancing information density, removing occlusions, and improving overall video quality.

Multi-view camera recording systems are deployed in various scenarios, including sports events \cite{staelens2015impact,hudctracker,chen2019learning,daniyal2010content}, office settings \cite{liu2001automating,doubek2004cinematographic} and open surgery \cite{bib1,saito2021camera,bib2,takatsuki2024construction,kato2023high,masuda2022novel,obayashi2023multi,fujii2022surgical,hachiuma2020deep}. Given their ability to capture extensive video footage, there is a growing need for automatic viewpoint switching or video summarization techniques to efficiently distill the essential information from the vast amount of data collected. In open surgery, Liu Xiang \cite{LiuxiangPaper} collaborated with medical experts to design a rule-based mechanism for assessing key entity detection to guide camera selection. Shimizu et al. \cite{bib2} developed a camera selection algorithm using image segmentation, trained via manual labeling and Dijkstra's algorithm. Hachiuma et al. \cite{hachiuma2020deep} introduced a fully supervised deep neural network that predicts the optimal camera view under expert guidance.
All the above methods have limitations. The first selects camera angles based only on the surgical area's size, ignoring video dynamics and potential occlusions as the scene changes over time. The second method relies solely on single-channel image features, lacking integration of multi-stream video characteristics. Thus, while effective in some cases, these methods require further optimization for accurate and flexible real-time camera switching in complex surgical environments.
Sarito et al. \cite{saito2021camera} proposed a different approach, training a model via self-supervision and using first-person videos to avoid complex manual annotations. Through transfer learning, they applied the model to shot selection, but this method requires entirely new video data. Their research adopted a semi-supervised approach, yet its performance lags behind the supervised algorithms reviewed in this paper.

This paper examines the temporal characteristics of occlusion in multi-channel videos captured by multi-viewpoint cameras mounted on shadowless lamps. Here, occlusion refers to instances where an object, such as medical instruments or a surgeon's hands, blocks the camera's view, resulting in certain angles where the target scene is partially or entirely obscured. Surgical videos record dynamic and continuous time sequences that consist of a series of interconnected steps and operations. Our method, by considering the temporal characteristics of these sequences, can better understand the correlation and sequence of occlusions in video frames \cite{tian2021weakly}. The method can more accurately select the optimal lens at each moment, providing a comprehensive and uninterrupted view of the surgical procedure.

The primary contributions of this paper are threefold: (1) We apply time-series prediction models to capture temporal data features, addressing the challenge of selecting cameras in surgical recordings to eliminate occlusion. (2)Latent semantic feature vectors are transformed into dense vector representations through feature embedding, reducing computational complexity and enhancing model efficiency. (3) We compare the performance of various time-series prediction model architectures and assess the impact of data structure transposition on model performance. This paper conducted a comprehensive evaluation of related methods using a dataset we created, demonstrating that our approach shows superior efficacy compared to similar methods.

\section{Related Work}
\subsection{Automatic camera switching from multi-viewpoint cameras}\label{cameras}
Multi-viewpoint camera switching algorithms are widely used in bioinformatics, sports events \cite{staelens2015impact}, traffic detection \cite{yang2023novel}, and video surveillance \cite{nguyen2003multiple}, among other fields. Liu Xiang et al. \cite{liu2021design} proposed a system that installed multiple cameras at various angles on a shadowless lamp to collect data from the surgical field. Their system assumes that at least one camera can capture the surgical target unobstructed. Shimizu et al. \cite{bib2} proposed a camera selection algorithm based on image segmentation, trained through manual labeling and Dijkstra optimization. They employed image segmentation \cite{li2013pixel} techniques, including color and texture-based division, to calculate the area of the surgical region. 
 Although commonly used, detecting the size of the crucial area may not be optimal for switching cameras based on the degree of occlusion in the surgical scene. Hachiuma et al. \cite{hachiuma2020deep} use a fully supervised convolutional neural network (CNN) that predicts the best-view camera by considering key factors such as the movement or posture of the doctor's hands and surgical tools. In their camera selection process, they considered the size of the surgical area and relied on human-annotated labels. 
 
 Saito et al. \cite{saito2021camera} introduced a camera selection method utilizing self-supervised learning to address occlusion issues in surgical recordings. This method leverages first-person perspective video from an eye tracker on the surgeon's head and footage from multiple cameras positioned under the operating light. Employing variational autoencoders (VAE) for self-supervised learning, the approach can automatically identify the optimal camera view without requiring manual labeling. While this approach enables unsupervised learning, it necessitates the acquisition of a substantial volume of new video data from head-mounted cameras worn by surgeons, and utilizing variational autoencoders for self-supervised learning might compromise the interpretability of the algorithm \cite{matsumoto2013digital}. Generally, consecutive frames have a strong correlation in addressing the issue of occlusion. However, in the research mentioned, the significance of temporal features is often overlooked in the exploration of multi-viewpoint camera switching tasks.

\begin{figure*}[h]
\centering
\includegraphics[width=0.7\textwidth]{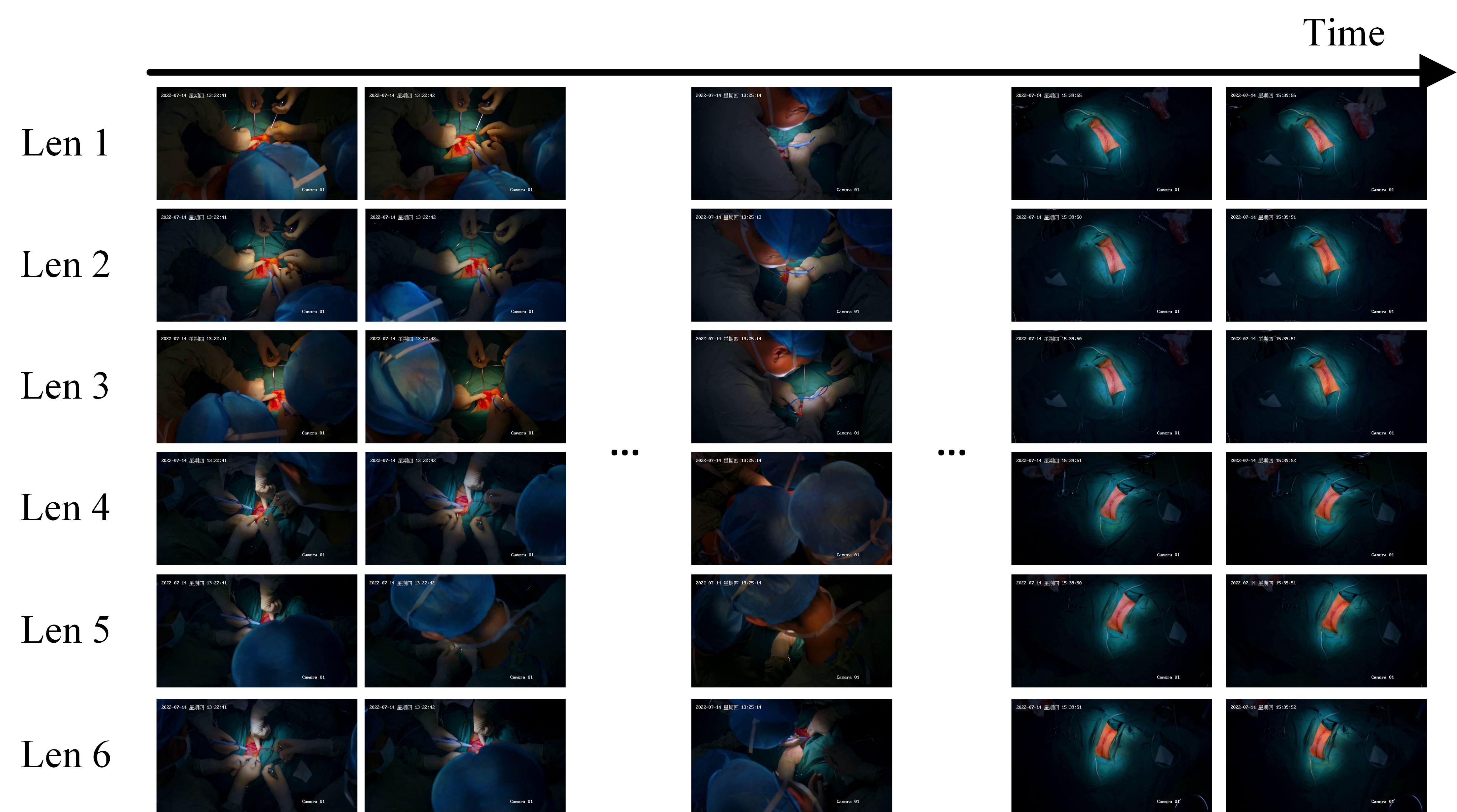}
\caption{
Multi-viewpoint cameras mounted on the shadowless surgical lamp allow the surgical procedure to be recorded simultaneously from six distinct perspectives. The shadowless lamp ensures consistent illumination, eliminating shadows in the surgical field and enabling each camera to capture critical procedural details precisely.}\label{fig1}
\end{figure*}

\subsection{Object Detection in Complex Surgical Scenarios}\label{Scenarios}
Object detection technology has found widespread use across various practical domains, including medicine. In emerging surgical areas like minimally invasive and robot-assisted surgeries, the integration of computer vision technology has reached a certain level of maturity \cite{wang2022visual}. However, research indicates that the development of this technology in open surgery still lags behind other fields \cite{zhang2020using}. With advancements in deep learning and neural networks, the analysis of surgical video data has become increasingly refined \cite{al2018monitoring}. Yet, compared to minimally invasive procedures, open surgery video data presents greater ambiguity and more interference factors \cite{saun2019video}, making the efficient collection and processing of this data crucial.

With the rise of deep neural networks, especially convolutional neural networks (CNNs), deep learning-based object detection has gained popularity. These methods achieve higher accuracy in complex surgical environments. Zhang et al. \cite{zhang2020using} proposed a CNN-based hand detection model that, combined with an object-tracking algorithm, enables precise hand detection and tracking. Basiev et al. \cite{basiev2022open} developed a method for classifying surgical tools using multi-view video data, addressing the issue of tool invisibility due to occlusions. Liu et al. \cite{bib1} introduced a YOLOv5-based approach to detect objects in surgical video frames. Fujii et al. \cite{fujii2022surgical} experimented with different pre-trained backbones to extract features and used various network structures, such as Faster R-CNN and RetinaNet. Goodman et al. \cite{goodman2021real} trained an AI model to analyze key elements of surgical procedures, using a multi-task model to generate surgical signatures and assess surgical skill through hand pose analysis. These studies demonstrate that, despite challenges such as occlusions, advancements in deep learning techniques have made significant progress in surgical video analysis.

\subsection{Deep learning for multivariate time-series forecasting}\label{multivariate time-series forecasting}
Time series forecasting involves predicting future trends in time series data. It has a greater demand for applications in fields, for instance, in electricity planning \cite{farnoosh2021deep}, transportation \cite{qu2019daily}, and financial strategic guidance \cite{lefrancois2020expectation}.

One of the core challenges in time series forecasting is modeling temporal variations, with many classical methods assuming that these variations follow predefined patterns \cite{hyndman2018forecasting,taylor2018forecasting}.  but as data complexity increases, many deep learning models, such as TCN \cite{bai2018empirical} and RNN \cite{gu2021efficiently}, have been developed for temporal modeling. TimesNet by Wu et al. \cite{wu2022timesnet} enhances forecasting accuracy by leveraging joint time-frequency modeling, which emphasizes the extraction of temporal and periodic features. 

Transformers\cite{vaswani2017attention} have demonstrated outstanding performance in time series forecasting with their attention mechanism effectively capturing dependencies between time points. Wu et al. \cite{wu2021autoformer} introduced the Autoformer model, which employs an Auto-Correlation mechanism to capture periodic dependencies and utilizes a deep decomposition architecture to extract seasonal and trend components from the input series. Informer by Zhou et al. \cite{zhou2021informer} enhances the efficiency of long-sequence forecasting through a sparse self-attention mechanism, while Zhang et al. \cite{zhang2023crossformer} introduced Crossformer, which improves the modeling of complex temporal patterns by capturing cross-domain dependencies. 

In this paper, We analyzed the multi-scale periodic characteristics of video-semantic fusion feature time series to capture their multi-periodic changes. 

\section{Method}\label{Method}
\subsection{Problem formulation}\label{formulation}
 In this paper, our objective is to predict a sequence of camera switching timing labels $ \bm{y} = [\bm y_{1},\bm y_{2},...\bm y_{T}]$, from synchronized video frames shot $\bm{I} = [\bm{I_{1}},\bm{I_{2}},...,\bm{I_{T}}]$, where $\bm{I_{t}} = [i_{t}^{1},i_{t}^{2},...,i_{t}^{N}]$, by $N$ cameras (in our experiments, $N$ = 6), $\bm y_{t}$ is a $N$-dimensional one-hot vector. Specifically, for each frame, we need to determine which camera provides the best unobstructed image. This problem can be simplified into an $N$-class classification problem, where each frame shot $\bm{I_{t}}$ is classified into one of the $N$ cameras. We use the softmax output to represent an N-dimensional one-hot vector for camera selection and choose the best camera based on a comparison of the output probabilities.

\begin{table*}[ht]
\centering
\begin{threeparttable}
\caption{The number of frames and chance rate for the training, validation, and test sets in our dataset.}
\label{datasets}
\begin{tabular*}{\linewidth}{@{\extracolsep{\fill}}lcccccc}  
\toprule
 & \multicolumn{2}{c}{Train} & \multicolumn{2}{c}{Validation} & \multicolumn{2}{c}{Test} \\
\cmidrule(lr){2-3} \cmidrule(lr){4-5} \cmidrule(lr){6-7}
Sequence & Frames & Chance Rate & Frames & Chance Rate & Frames & Chance Rate \\
\midrule
Surgery 1 & 5,308 & 0.515 & 758 & 0.489 & 1,516 & 0.531 \\
Surgery 2 & 4,010 & 0.489 & 573 & 0.483 & 1,145 & 0.497 \\
Surgery 3 & 2,400 & 0.512 & 343 & 0.609 & 686 & 0.513 \\
Surgery 4 & 5,011 & 0.476 & 716 & 0.605 & 1,432 & 0.607 \\
Surgery 5 & 3,405 & 0.336 & 486 & 0.534 & 973 & 0.554 \\
\midrule
Total & 20,134 & 0.469 & 2,876 & 0.539 & 5,752 & 0.545 \\
\bottomrule
\end{tabular*}
\vspace{+1mm} 
\begin{tablenotes}
\footnotesize
\raggedright 
\item[]  Note: The dataset was randomly split into training (70\%), validation (10\%), and test (20\%) sets, using a fixed random seed to ensure balanced data distribution. This approach ensures sufficient training data while providing reliable validation and test sets, allowing for accurate evaluation of the model's generalization and reducing potential bias from uneven splits.
\end{tablenotes}
\end{threeparttable}
\end{table*}

\subsection{Datasets}
Since no publicly available datasets exist for multi-camera recordings in open surgery, we developed our dataset using the method proposed by Liu et al. [4] in their paper. Our dataset consists of recordings from multiple angles captured by cameras mounted on surgical lights during five distinct thyroidectomy procedures, with a frame rate of 30 frames per second. After anonymizing the data, we synchronized the frames to ensure alignment across all multi-channel camera videos. Given the minimal scene changes and the extended duration of open surgery videos, we selected keyframes for annotation at 1-second intervals.

After completing data preprocessing, experienced thyroidectomy surgeons manually annotated the dataset to identify the optimal camera views. The annotation process was designed to minimize occlusions in the selected images and reduce interference with semantic information extraction. We developed custom annotation software specifically for labeling camera selection data. To ensure annotation accuracy, each group independently annotated randomly shuffled multi-angle image pairs. Any discrepancies between annotations were reviewed and resolved. This approach resulted in the development of a high-quality dataset suitable for practical training and testing. Table~\ref{datasets} below presents the number of frames and chance rate for the training, validation, and test sets in our dataset.

\begin{figure}[ht]
\centering
\includegraphics[width=0.5\textwidth]{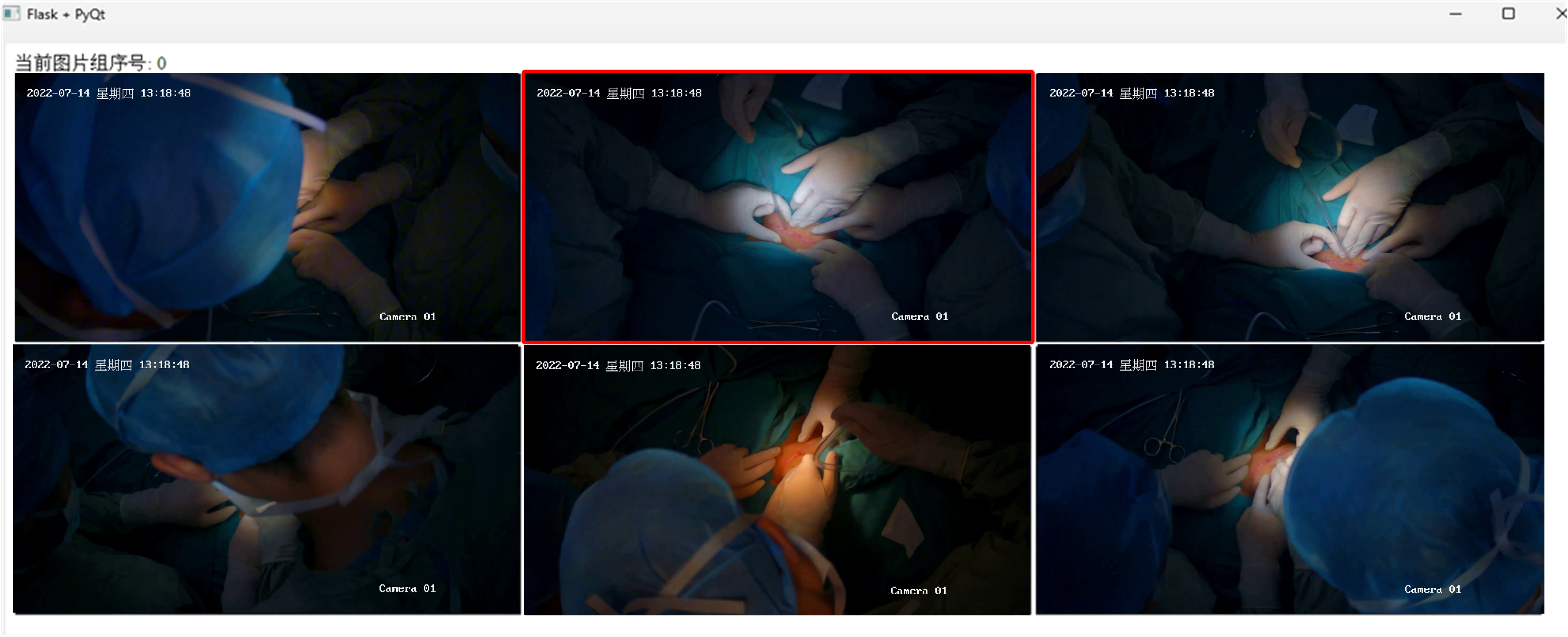}
\caption{Annotation software interface: simultaneously displaying images from six different camera angles at the same time, allowing the annotator to select the best angle for annotation by clicking on the image.}\label{fig2}
\end{figure}

\subsection{Network Architecture}
\begin{figure*}[ht]
\centering
\includegraphics[width=1\textwidth]{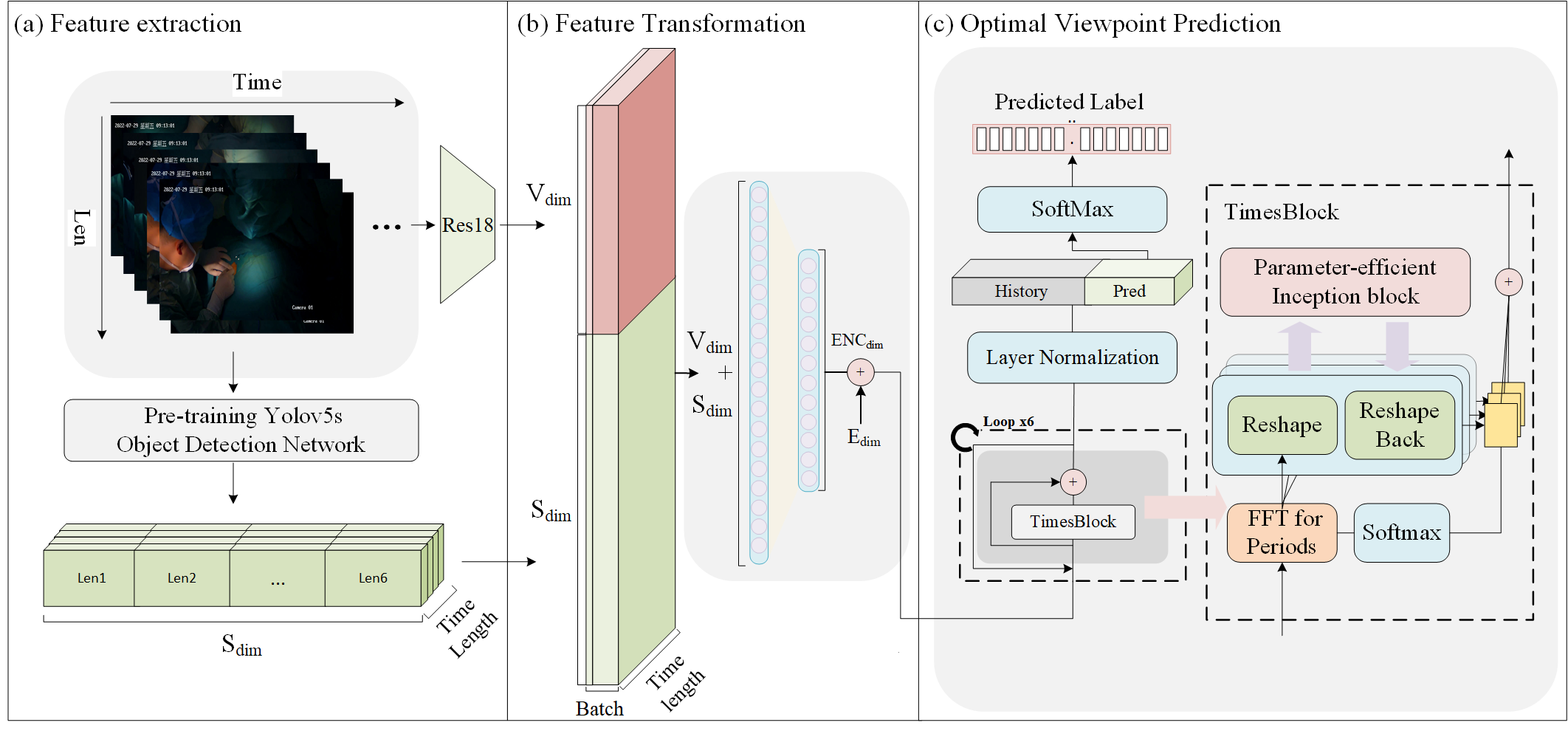}
\caption{The overall architecture of an end-to-end time-series prediction of multi-angle camera selection in open surgery: (a) Feature Extraction: A pre-trained ResNet-18 model is employed to extract visual features, while semantic features are extracted using the YOLOv5s model. These features are then integrated as inputs to the temporal prediction network. (b) Feature Transformation: Dimensionality reduction is performed on the high-dimensional feature vectors using a linear embedding layer, with temporal information incorporated to enhance the model's ability to capture sequential dependencies. (c) Optimal Viewpoint Prediction: The TimesBlock modules process the time-series data, with the Softmax classifier generating a probability distribution over possible camera labels, from which the optimal label is selected.}\label{fig111}
\end{figure*}

The proposed network architecture is composed of three primary components: feature extraction, feature transformation, and optimal viewpoint prediction. 
In the feature extraction stage, a pre-trained ResNet-18 model is employed to extract visual features $\bm{v_{t}^{n}}$ from images $i_{t}^{n}$, at each step. These are subsequently concatenated to form a comprehensive image feature representation. 

Semantic features are extracted using the object detection method developed by Liu et al \cite{bib1}. The method was adapted and fine-tuned using a YOLOv5s model pre-trained on their thyroidectomy dataset. The extracted semantic features $\bm{s_{t}^{n}}$  for each frame include the number of detected objects, their coordinates, bounding box dimensions (length and width), and bounding box area for n perspectives of images. The input for the temporal prediction network is constructed by integrating visual and semantic features.
\begin{equation}
V_{\text{dim}} = \{ V_{1}, V_{2}, \dots, V_{T} \}, \quad V_{t} = v_{t}^{1} \oplus v_{t}^{2} \oplus \dots \oplus v_{t}^{N}
\end{equation}
\begin{equation}
S_{\text{dim}} = \{ S_{1}, S_{2}, \dots, S_{T} \} , \quad S_{t} = s_{t}^{1} \oplus s_{t}^{2} \oplus \dots \oplus s_{t}^{N}
\end{equation}

\begin{table}[htbp]
\centering
\caption{Detected Objects and Simplified Descriptions}
\begin{tabular}{ll}
\toprule
\textbf{Items detected} & \textbf{Simplified explanation} \\
\midrule
aspirator & Suctions fluids from surgical site. \\
bistoury & Small knife for precise cutting. \\
detector & Identifies specific objects or devices. \\
drainage tube & Removes excess fluids from body. \\
electrotome & Cauterizes tissue using electrical current. \\
gauze & Absorbs fluids and covers wounds. \\
glue & Adhesive used to close incisions. \\
hand & Surgeon’s hand involved in procedure. \\
head & Surgeon’s head during the operation. \\
hemostat & Clamps blood vessels to stop bleeding. \\
injector & Device for administering injections. \\
nesis & Surgical thread for suturing wounds. \\
porteaiguille & Needle holder for suturing. \\
sterile patches & Sterile dressings for wound protection. \\
thyroid retractor & Retracts tissue for thyroid exposure. \\
thyroid retractor back & Retracts tissue behind the thyroid. \\
thyroid retractor front & Retracts tissue in front of thyroid. \\
thyroid tissue & Tissue from the thyroid gland. \\
tissue scissors & Scissors for precise tissue cutting. \\
towel forceps & Grasp towels or dressings. \\
treatment bowl & Holds fluids or instruments during surgery. \\
tweezer & Precision tool for gripping small objects. \\
wound & Surgical incision site. \\
\bottomrule
\end{tabular}
\end{table}

In the feature transformation phase, we converge visual and semantic features into a unified high-dimensional vector. To mitigate the computational load associated with such high-dimensional data, we employ a linear embedding layer for dimensionality reduction, skillfully mapping our feature space to a more tractable, lower-dimensional representation. This process not only compacts the feature vector but also enhances the model's efficiency and predictive accuracy. By incorporating temporal information, represented as $E_{t}$, into the feature vector, our model gains the ability to capture the sequential dependencies characteristic of time-series data. The essence of this process is captured in the following equation:
\begin{equation}
    Enc_{\text{dim}} = \text{Dropout}(\sigma (W \cdot (V_{\text{dim}} \oplus S_{\text{dim}}) + b)) + E_t
\end{equation}

Here, $Enc_{\text{dim}}$ represents the resultant feature vector after dimensionality reduction.
The high-dimensional vector $V_{\text{dim}} \oplus S_{\text{dim}}$ transformed by the weight matrix $W$ and bias $b$. The activation function $\sigma $ introduces non-linearity, and Dropout helps prevent overfitting by randomly setting a fraction of the output units to zero during training. 

To achieve optimal Viewpoint prediction, the model employs multiple timesBlock modules combined with residual connections to handle long-term, multivariate time-series data effectively. The residual connections not only mitigate the vanishing gradient problem but also enhance the model’s ability to capture long-term dependencies between time steps, thereby improving learning efficiency and model stability. After layer normalization, the processed temporal feature vectors $Z$ are fed into a softmax layer for classification, which computes the probability distribution $P$ over camera labels as follows:

\begin{equation}
P(y = n \mid Z) = \frac{\exp(W_n^\top Z + b_n)}{\sum_{j=1}^n \exp(W_j^\top Z + b_j)}
\end{equation}

where $W_n$ and $b_n$ are the weights and biases for the $n$-th camera label, and $P(y = n)$ is the probability of selecting camera label $n$. The model is optimized during training using weighted cross-entropy loss, and in inference, the camera label with the highest probability is selected as the final output.

\subsection{Network training}\label{training}
In this experiment, we applied our model to long-term time series forecasting, training it on a preprocessed camera dataset with an input sequence length of 12 and a prediction length of 6.$in 2$ The training process used the CrossEntropyLoss function for optimization. We carefully selected hyperparameters, including input feature dimensions, batch size, learning rate, and the number of layers, to enhance training performance and generalization. We used an NVIDIA A100 GPU to expedite the training process. The model was trained with a batch size of 8 across 10 epochs. An early stopping mechanism (patience=5) halted training if no significant improvement was observed over 10 consecutive epochs. To prevent overfitting, a dropout rate of 0.3 was applied. A learning rate adjustment strategy dynamically optimized parameters when validation performance plateaued. Ultimately, the model demonstrated excellent performance in the camera data time series forecasting task through multiple experiments and hyperparameter tuning, significantly improving prediction accuracy.

To address the issue of class imbalance in the dataset, we use weighted cross-entropy to introduce weights, assigning higher weights to the minority classes. This encourages the model to pay more attention to these underrepresented classes, reducing the tendency to favor the majority classes during prediction.

\begin{equation}
L = - \frac{1}{N} \sum_{i=1}^{N} \sum_{c=1}^{C} w_c \cdot y_{ic} \log(\hat{y}_{ic})
\end{equation}

\subsection{Analysis of evaluation methods}\label{Evaluation}
To evaluate the performance of the model, we designed two evaluation approaches: Sequence-Out and Surgery-Out, to comprehensively analyze the model's camera selection performance in multi-camera surgical scenarios.

In the Sequence-Out evaluation, the model was trained with data from all surgery types, but the test sequences differed from the training sequences. Although the model was familiar with the surgery types during training, it had never encountered the specific test sequences, requiring it to select the best camera based on unknown sequences.

The Surgery-Out evaluation posed a greater challenge. The model was trained on multiple surgery sequences, but the surgery video used in testing had never appeared during training. This setting increased the difficulty as the model had to handle not only new sequences but also completely unfamiliar surgery videos. In the Sequence-Out setting, the test data shared the same surgery types as the training data, but the test sequences were new, assessing the model’s ability to handle new sequences within known surgery types. 

In contrast, the Surgery-Out setting required the model to deal with entirely new sequences and surgery types. This stricter evaluation tested the model’s adaptability in various scenarios, where it needed to select the best camera even in completely unfamiliar environments. Overall, the Sequence-Out evaluation focuses more on assessing the model’s generalization ability in known conditions, while the Surgery-Out evaluation tests the model’s performance when dealing with unknown surgery videos.

\section{Results}\label{Results}
\subsection{Evaluation of Sequence-Out and Surgery-Out}\label{Evaluation1}
In the Sequence-Out evaluation, all five surgical video sequences were used for training, while validation and testing were performed on a specific single surgical video sequence. When selecting a particular sequence for model evaluation, we allocated 70\% of the sequence to the training set, 10\% to the validation set, and the remaining 20\% to the test set. In the Surgery-Out evaluation, the model was trained on four surgical video sequences, with the remaining sequence not included in the training set reserved for validation and testing. For this sequence, we selected 20\% for the test set and 10\% for the validation set. Although all five different surgical sequences belong to the same type of surgery, there are significant differences in the surgical processes, success rates, lighting conditions, and durations, leading to notable variations in frame conditions. Therefore, selecting the camera in such a setting is complex. Table~\ref{tabreulst1} presents the validation results. We used two different pre-trained models to extract semantic and image features. By concatenating the mixed feature inputs and utilizing a temporal prediction neural network, our approach outperformed the baseline in terms of accuracy. Furthermore, our method was compared with other supervised learning algorithms (e.g., Shimizu et al., Hachiuma et al.), demonstrating superior performance on our dataset, even with a longer prediction sequence compared to their models.

In this section, we primarily compare our method with two previous camera-switching algorithms, while also setting a baseline approach that does not utilize the semantic information collected through our pre-trained object detection model.

\textbf{Shimizu \textit{et al.} \cite{bib2}}
: They proposed a supervised learning algorithm aimed at selecting the camera that maximizes the surgical field area. The method calculates the surgical region's area using image segmentation \cite{li2013pixel} and optimizes the camera sequence through the Dijkstra algorithm. As the code and model were not publicly available, we recreated their approach based on the paper's description. Despite adhering closely to the method outlined, minor discrepancies may still exist.

\textbf{Hachiuma \textit{et al.}  \cite{hachiuma2020deep}}
: A network was developed to predict the optimal viewpoint, with training conducted on surgical lump video frames annotated by experts. As the original code and model were inaccessible, their method was reconstructed based on the paper’s description. Although we strictly adhered to the outlined methodology, minor implementation discrepancies may still exist.

\textbf{Hachiuma \textit{et al.}  with Semantic Features}
: A fully supervised camera selection network designed to directly predict the optimal-view camera. Based on the original algorithm, we incorporated semantic information data obtained from a pre-trained object detection model into the model's input.

\textbf{Ours w/o Semantic Features}: This network, designed for fully supervised camera selection, aims to predict the optimal camera view by leveraging only visual data. Here, we modified the original algorithm by excluding the semantic information extracted from the pre-trained object detection model, leaving other components of the input unchanged.

\textbf{Ours w/o Video Features}: A variation of the fully supervised camera selection network, this model predicts the best camera view based solely on semantic data. Adhering to the original framework, we removed video data obtained from the pre-trained ResNet-18 model, isolating the impact of video features on camera selection.

\textbf{Ours}: This fully supervised camera selection network directly predicts the optimal view by integrating both visual and semantic features. Leveraging pre-trained ResNet-18 and YOLOv5s models, it concatenates these features into a high-dimensional vector, which is then passed through a linear embedding layer for dimensionality reduction. The processed multivariate time-series data enables the model to generate a sequence of optimal camera angles for each frame, maintaining the original algorithm's structure while maximizing feature use.

\begin{table*}[h]
\caption{Accuracy evaluation of camera selection performance in Sequence-Out and Surgery-Out settings.}
\label{tabreulst1}
\centering
\begin{threeparttable}
\begin{tabular*}{\textwidth}{@{\extracolsep{\fill}}lcccccc}
\toprule
& \multicolumn{6}{c}{Sequence-Out}\\\cmidrule{2-7}
\raisebox{2ex}[0pt]{Methods} & Surgery 1 & Surgery 2 & Surgery 3 & Surgery 4 & Surgery 5 & Average \\
\midrule
Shimizu et al.   & 0.608 & 0.715 & 0.758 & 0.689 & 0.716 & 0.701\\
\midrule
Hachiuma et al. & 0.797 & 0.821 & 0.835 & 0.823 & 0.826 & 0.820\\
Hachiuma et al. w/o Semantic Features & 0.807 & 0.756 & 0.844  & 0.826 & 0.822 & 0.811\\
\midrule
Ours w/o Video Features  & 0.802 & 0.786  & 0.820 & 0.807 & 0.832 & 0.809\\
Ours w/o Semantic Features  & 0.863 & $\mathbf{0.871}$  & 0.880  & 0.891 & 0.873 & 0.875\\
Ours  & $\mathbf{0.919}$ & 0.869 & $\mathbf{0.923}$ & $\mathbf{0.920}$ & $\mathbf{0.925}$ &  $\mathbf{0.911}$\\
\midrule
& \multicolumn{6}{c}{Surgery-Out}\\\cmidrule{2-7}
\raisebox{2ex}[0pt]{Methods} & Surgery 1 & Surgery 2 & Surgery 3 & Surgery 4 & Surgery 5 & Average \\
\midrule
Shimizu et al.   & 0.602  & 0.572 & 0.589 & 0.670 & 0.656 & 0.618\\
\midrule
Hachiuma et al.  & 0.798 & 0.794 & 0.808 & 0.783 & 0.802 & 0.797\\
Hachiuma et al. w/o Semantic Features & 0.772 & 0.773  & 0.785  & 0.808 & 0.802 & 0.788\\
\midrule
Ours w/o Video Features  & 0.659 & 0.658  & 0.648 & 0.694 & 0.691 & 0.670\\
Ours w/o Semantic Features  & $\mathbf{0.889}$  & 0.890 & $\mathbf{0.924}$ & 0.881 & $\mathbf{0.893}$ & $\mathbf{0.895}$\\
Ours  & 0.867 & $\mathbf{0.891}$ & 0.880 & $\mathbf{0.909}$ &  $\mathbf{0.893}$ & 0.888\\
\bottomrule
\end{tabular*}
\vspace{+1mm} 
\begin{tablenotes}
\footnotesize
\raggedright
\item[] Note: In the Sequence-Out and Surgery-Out settings, we evaluated the performance of camera selection using prediction accuracy, where higher accuracy indicates better model performance. This method utilized 128-dimensional embedding feature vectors as input.
\end{tablenotes}
\end{threeparttable}
\end{table*}

\begin{table*}[htbp]
\caption{Comparative performance of time series prediction algorithms with varying input and prediction lengths.}\label{tabrelust2}
\begin{tabular*}{\textwidth}{@{\extracolsep\fill}lcccccccc}
\toprule%
& \multicolumn{2}{c}{Sequence}& \multicolumn{6}{c}{Sequence-Out }\\\cmidrule{2-9}%
\raisebox{2ex}[0pt]{Methods} & input & pred & surgery 1 & surgery 2 & surgery 3 & surgery 4 & surgery 5 & Average \\
\midrule
\multirow{3}{*}{Autoformer} & 12 & 6 & 0.872& 0.834 & 0.895 & 0.870 & 0.900 & 0.874 \\
& 60 & 30  & 0.892 & 0.882  & 0.904 & 0.898 & 0.899 & $\mathbf{0.895}$ \\
& 120 & 60  & 0.887 & 0.877 & 0.900 & 0.911 & 0.893 & $\mathbf{0.894}$  \\
\midrule

\multirow{3}{*}{Informer}  & 12 & 6  & 0.752 & 0.667 & 0.799 & 0.798 & 0.816 & 0.766\\
 & 60 & 30 & 0.726 & 0.640 & 0.798 & 0.780 & 0.801 & 0.749\\
 & 120 & 60 & 0.763 & 0.575 & 0.792 & 0.792 & 0.802 & 0.745 \\
\midrule
\multirow{3}{*}{Crossformer}  & 12 & 6 & 0.812 & 0.707 & 0.707 & 0.804 & 0.826 & 0.771\\
& 60 & 30 & 0.743 & 0.593 & 0.809 & 0.781 & 0.807 & 0.747\\
& 120 & 60 & 0.709 & 0.716 & 0.795 & 0.847& 0.858 & 0.785 \\
\midrule
\multirow{3}{*}{Ours}   & 12 & 6  & 0.919& 0.869& 0.923 & 0.920& 0.925 &  $\mathbf{0.911}$\\
& 60 & 30 & 0.876 & 0.821 & 0.898 & 0.898 & 0.901 & 0.879 \\
& 120 & 60 & 0.837 & 0.745 & 0.858 & 0.877 & 0.870 & 0.837 \\
\bottomrule

& \multicolumn{2}{c}{Sequence}& \multicolumn{5}{c}{Surgery-Out }\\\cmidrule{2-9}%
\raisebox{2ex}[0pt]{Methods} & input & pred & surgery 1 & surgery 2 & surgery 3 & surgery 4 & surgery 5 & Average\\
\midrule
\multirow{3}{*}{Autoformer} & 12 & 6 & 0.878 & 0.851 & 0.871 & 0.870 & 0.875 &  0.869  \\
& 60 & 30 & 0.893 & 0.887 & 0.864  & 0.862 & 0.865 & $\mathbf{0.874}$ \\
& 120 & 60 & 0.939 & 0.897 & 0.884 & 0.896 & 0.860 & $\mathbf{0.895}$ \\
\midrule

\multirow{3}{*}{Informer} & 12 & 6 & 0.690 & 0.673 & 0.742 & 0.777 & 0.754 &  0.727  \\
 & 60 & 30 & 0.654 & 0.597 & 0.620 & 0.710 & 0.623 & 0.641  \\
 & 120 & 60 & 0.671 & 0.590  & 0.615 & 0.751 & 0.708 &  0.667\\
\midrule
\multirow{3}{*}{Crossformer}  & 12 & 6 & 0.702 & 0.761 & 0.799 & 0.791& 0.763 &  0.763\\
& 60 & 30 & 0.703 & 0.708 & 0.627 & 0.722 & 0.775 & 0.707 \\
& 120 & 60 & 0.770 & 0.678 & 0.773 & 0.817 & 0.733 & 0.754\\
\midrule
\multirow{3}{*}{Ours}   & 12 & 6 & 0.867 & 0.891 & 0.880 & 0.909 &  0.893 & $\mathbf{0.888}$\\
& 60 & 30 & 0.832 & 0.868 & 0.847 & 0.855 & 0.868 & 0.854\\
& 120 & 60 & 0.837 & 0.745 & 0.858 & 0.877 & 0.870 & 0.837\\
\bottomrule
\end{tabular*}
\end{table*}

\subsection{Evaluation of Time-series-forcasting}\label{Evaluation2}

In this section, we conducted a series of comparative experiments aimed at evaluating and comparing the performance of various algorithms in the field of time series forecasting. To ensure that our assessment is comprehensive and accurate, we designed a set of experiments that included a variety of input lengths and prediction lengths. Input length refers to the amount of historical data that the model receives, while prediction length refers to the number of future time steps the model needs to predict. By varying these parameters, we can better understand the performance of different algorithms under different conditions.

We selected several advanced network frameworks for these experiments, including Autoformer, Informer, and Crossformer. These frameworks are the latest technologies proposed in the field of time series forecasting, each with its unique advantages and characteristics:

    \textbf{Autoformer}\cite{wu2021autoformer}: This is a variant of the self-attention mechanism that can automatically learn the temporal dependencies in the data without explicit recursive or convolutional structures.
    
      \textbf{Informer}\cite{zhou2021informer}: This is a Transformer-based model specifically designed to handle long-sequence forecasting problems. It improves computational efficiency through a novel attention mechanism.
      
    \textbf{Crossformer}\cite{zhang2023crossformer}: This is a model that combines cross-attention mechanisms, enabling it to capture the interrelationships between different time series.

In the experiments, we iterated each algorithm multiple times to ensure the stability and reliability of the results. The experimental results were meticulously recorded and presented in Table~\ref{tabrelust2}. These results include not only the accuracy of the predictions but may also encompass other important metrics such as computational efficiency and the model's generalization capabilities.

\section*{Conclusion}

In this paper, we created a dataset for the task of selecting the best-view camera and used time series prediction models to solve the task of selecting the optimal camera from multiple open surgery videos. Our approach extracts latent semantic feature vectors and video feature vectors from images using pre-trained object recognition and image feature extraction models. By applying feature embedding, we transform sparse feature data into dense latent feature vector representations, reducing computational complexity and improving efficiency. Additionally, we compared the performance of time series prediction models with different frameworks on this task, and extensively evaluated methods proposed by other researchers using the dataset we created. Our approach demonstrated promising effectiveness compared to other comparable methods. Through these contributions, our research provides an effective solution for selecting the best camera view in surgical videos.

\section*{Discussion}
\setcounter{subsection}{0}  
\subsection{Performance and model insights}
The results of this study demonstrate that using dense latent feature vector representations through embedding greatly improves computational efficiency while preserving a high level of accuracy. This indicates that the proposed time series prediction approach is robust and adaptable to varying testing conditions. The method’s ability to capture temporal patterns and dependencies effectively highlights its potential for practical applications in optimal camera view selection within open surgery environments. The observed improvements validate the model's suitability for real-time implementation in scenarios where quick and accurate viewpoint decisions are critical.

\subsection{Limitations and potential biases}
Despite its strong performance, the model may face limitations when applied to surgical types not included in the current dataset, potentially impacting its generalizability across other open surgical procedures. Additionally, biases inherent in the pre-trained models used for feature extraction—such as training data limitations or object recognition biases—may affect representation accuracy, which we aim to address in future iterations. Sustaining performance in longer sequences is challenging due to the potential accumulation of prediction errors, which can affect consistency in extended surgical procedures.

\subsection{Future research directions}
Future work will focus on adapting this model for real-time applications within operating rooms, where minimizing latency is crucial for supporting intraoperative decision-making. Additionally, we plan to integrate multimodal data sources, such as audio and physiological signals, to increase the model’s adaptability and provide richer contextual information for viewpoint decisions. This direction aims to make the camera view selection system more responsive and applicable in diverse and complex surgical settings.

\section*{Acknowledgment}

We would like to express our sincere gratitude to everyone who provided invaluable assistance in creating the dataset used for the best-view camera selection task in this paper. Your contributions were essential to the success of this work.

\section*{References}

\bibliographystyle{IEEEtran}
\bibliography{bibliography}

\end{document}